\documentclass[10pt,twocolumn,letterpaper]{article}

\usepackage{iccv}
\usepackage{times}
\usepackage{epsfig}
\usepackage{graphicx}
\usepackage{amsmath}
\usepackage{amssymb}

\usepackage{appendix}
\usepackage{pdfpages}

\usepackage{multirow}
\usepackage{float}
\usepackage[belowskip=-0pt,aboveskip=0pt]{caption}
\usepackage{bbding}
\usepackage{caption}
\DeclareCaptionFont{9pt}{\fontsize{9pt}{10pt}\selectfont}
\usepackage[marginal]{footmisc}

\usepackage[colorlinks,linkcolor=red]{hyperref}

\iccvfinalcopy 

\def\iccvPaperID{****} 

\ificcvfinal\fi

\begin{document}

\makeatletter
\renewcommand*{\@fnsymbol}[1]{\ensuremath{\ifcase#1\or \dagger\or \dagger\or \ddagger\or
    \mathsection\or \mathparagraph\or \|\or **\or \dagger\dagger
    \or \ddagger\ddagger \else\@ctrerr\fi}}
\makeatother

\title{RankSRGAN: Generative Adversarial Networks with Ranker\\ for Image Super-Resolution}

\author{Wenlong Zhang{\textsuperscript{1}} \qquad Yihao Liu{\textsuperscript{1,2}} \qquad  Chao Dong{\textsuperscript{1,\thanks{Corresponding author (e-mail: chao.dong@siat.ac.cn) }}} \qquad Yu Qiao{\textsuperscript{1}}\\
\textsuperscript{1}ShenZhen Key Lab of Computer Vision and Pattern Recognition, SIAT-SenseTime Joint Lab,\\
 Shenzhen Institutes of Advanced Technology, Chinese Academy of Sciences, China\\
\textsuperscript{2}University of Chinese Academy of Sciences \\
{\tt\small \{wl.zhang1, yh.liu4, chao.dong, yu.qiao\}@siat.ac.cn}
}


\maketitle
\ificcvfinal\thispagestyle{empty}\fi

\begin{abstract}
Generative Adversarial Networks (GAN) have demonstrated the potential to recover realistic details for single image super-resolution (SISR). To further improve the visual quality of super-resolved results, PIRM2018-SR Challenge employed perceptual metrics to assess the perceptual quality, such as PI, NIQE, and Ma. However, existing methods cannot directly optimize these indifferentiable perceptual metrics, which are shown to be highly correlated with human ratings. To address the problem, we propose Super-Resolution Generative Adversarial Networks with Ranker (RankSRGAN) to optimize generator in the direction of perceptual metrics. Specifically, we first train a Ranker which can learn the behavior of perceptual metrics and then introduce a novel rank-content loss to optimize the perceptual quality. The most appealing part is that the proposed method can combine the strengths of different SR methods to generate better results. Extensive experiments show that RankSRGAN achieves visually pleasing results and reaches state-of-the-art performance in perceptual metrics. Project page: \url{https://wenlongzhang0724.github.io/Projects/RankSRGAN}

\end{abstract}


\section{Introduction}

Single image super resolution aims at reconstructing/generating a high-resolution (HR) image from a low-resolution (LR) observation. Thanks to the strong learning capability, Convolutional Neural Networks (CNNs) have demonstrated superior performance   \cite{dong2014learning,lim2017enhanced,zhang2018residual} to the conventional example-based \cite{zhang2012single} and interpolation-based \cite{zhang2006edge} algorithms. Recent CNN-based methods can be divided into two groups. The first one regards SR as a reconstruction problem and adopts MSE as the loss function to achieve high PSNR values. However, due to the conflict between the reconstruction accuracy and visual quality, they tend to produce overly smoothed/sharpened images. To favor better visual quality, the second group casts SR as an image generation problem \cite{ledig2017photo}. By incorporating the perceptual loss \cite{bruna2015super,johnson2016perceptual} and adversarial learning \cite{ledig2017photo}, these perceptual SR methods have potential to generate realistic textures and details, thus attracted increasing attention in recent years. 

\begin{figure}[t]
\setlength{\abovecaptionskip}{-0.3cm}
\setlength{\belowcaptionskip}{-0.24cm}
\begin{center}
	\includegraphics[width=0.92\linewidth]{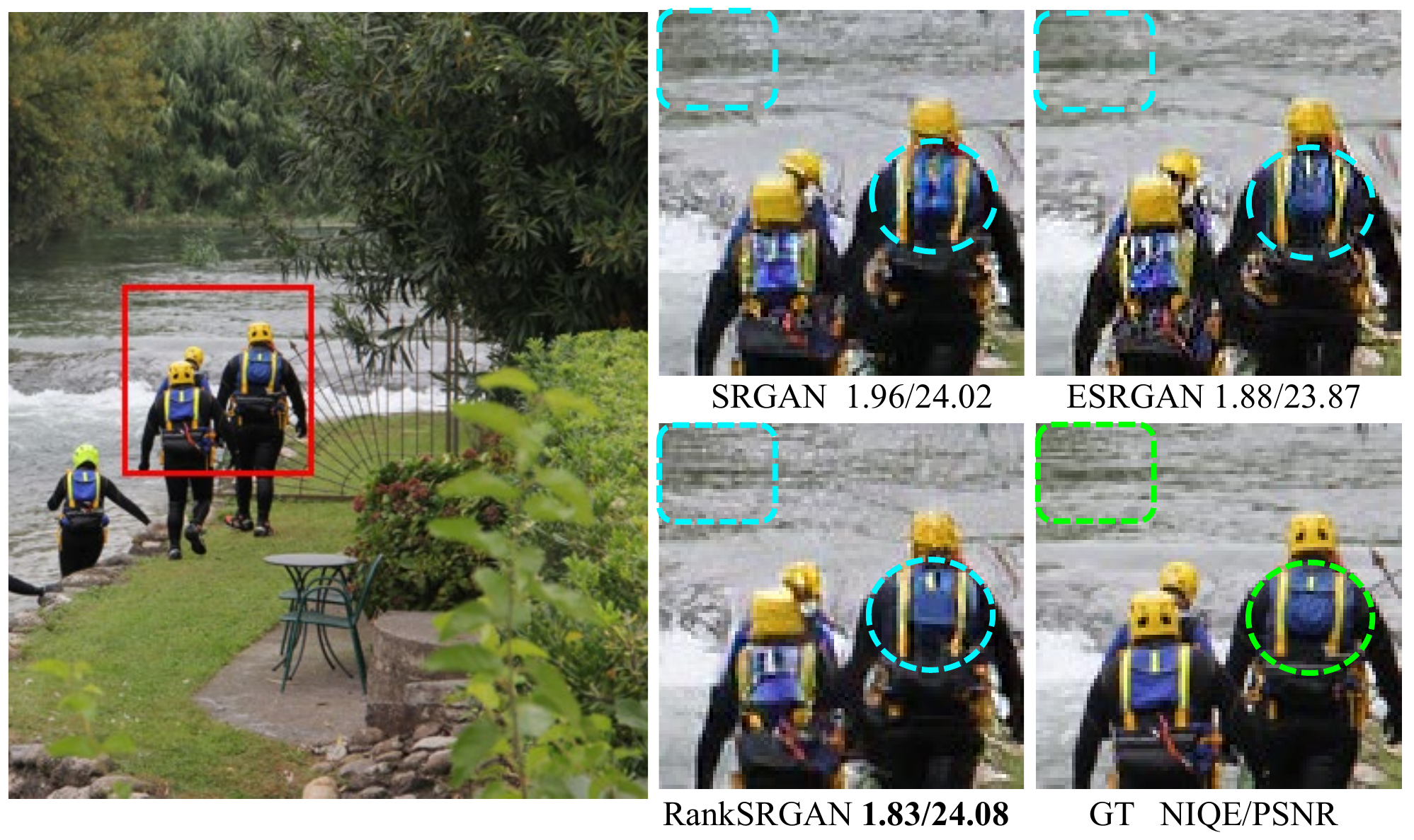} 

\end{center}
   \caption{The comparison of RankSRGAN and the state-of-the-art perceptual SR methods on $\times$4. NIQE: lower is better. PSNR: higher is better.}
\label{fig:1}
\label{fig:onecol}
\vskip -0.5cm
\end{figure}
The most challenging problem faced with perceptual SR methods is the evaluation. Most related works resort to user study for subjectively evaluating the visual quality \cite{blau20182018, wang2018recovering}. However, without an objective metric like PSNR/SSIM, it is hard to compare different algorithms on a fair platform, which largely prevents them from rapid development. To address this issue, a number of no-reference image quality assessment (NR-IQA) metrics are proposed, and some of them are proven to be highly correlated with human ratings \cite{blau20182018}, such as NIQE \cite{mittal2013making} (correlation 0.76) and PI \cite{blau20182018} (correlation 0.83). Specially, the PIRM2018-SR challenge \cite{blau20182018} introduced the PI metric as perceptual criteria and successfully ranked the entries. Nevertheless, most of these NR-IQA metrics are not differentiable (e.g., they include hand-crafted feature extraction or statistic regression operation), making them infeasible to serve as loss functions. Without considering NR-IQA metrics in optimization, existing perceptual SR methods could not show stable performance in the orientation of objective perceptual criteria.

To overcome this obstacle, we propose a general and differentiable model -- Ranker, which can mimic any NR-IQA metric and provide a clear goal (as loss function) for optimizing perceptual quality. Specifically, Ranker is a Siamese CNN that simulates the behavior of the perceptual metric by learning to rank approach \cite{burges2005learning}. Notably, as NR-IQA metrics have various dynamic ranges, Ranker learns their output ranking orders instead of absolute values. Just like in the real world, people tend to rank the quality of images rather than give a specific value. We equip Ranker with the standard SRGAN model and form a new perceptual SR framework -- RankSRGAN (Super-Resolution Generative Adversarial Networks with Ranker). In addition to SRGAN, the proposed framework has a rank-content loss using a well-trained Ranker to measure the output image quality. Then the SR model can be stably optimized in the orientation of specific perceptual metrics.

To train the proposed Ranker, we prepare another training dataset by labeling the outputs of different SR algorithms. Then the Ranker, with a Siamese-like architecture, could learn these ranking orders with high accuracy. The effectiveness of the Ranker is largely determined by the selected SR algorithms. To achieve the best performance, we adopt two state-of-the-art perceptual SR models -- SRGAN \cite{ledig2017photo} and ESRGAN \cite{Wang_2018_ECCV_Workshops}. As the champion of PIRM2018-SR challenge \cite{blau20182018}, ESRGAN is superior to SRGAN on average scores, but can not outperform SRGAN on all test images. When evaluating with NIQE \cite{mittal2013making}, we obtain mixed orders for these two methods. Then the Ranker will favor different algorithms on different images, rather than simply classifying an image into a binary class (SRGAN/ESRGAN). After adopting the rank-content loss, the generative network will output results with higher ranking scores. In other words, the learned SR model could combine the better parts of SRGAN and ESRGAN, and achieve superior performance both in perceptual metric and visual quality. Figure \ref{fig:1} shows an example of RankSRGAN, which fuses the imagery effects of SRGAN and ESRGAN and obtains better NIQE score. 

We have done comprehensive ablation studies to further validate the effectiveness of the proposed method. First, we distinguish our Ranker from the regression/classification network that could also mimic the perceptual metric. Then, we train and test RankSRGAN with several perceptual metrics (i.e. NIQE \cite{mittal2013making}, Ma \cite{ma2017learning}, PI \cite{blau20182018}). We further show that adopting different SR algorithms to build the dataset achieves different performance. Besides, we have also investigated the effect of different loss designs and combinations. With proper formulation, our method can clearly surpass ESRGAN and achieve state-of-the-art performance. 

In summary, the contributions of this paper are three-fold.
(1) We propose a general perceptual SR framework -- RankSRGAN that can optimize generator in the direction of indifferentiable perceptual metrics and achieve the state-of-the-art performance. 
(2) We, for the first time, utilize results of other SR methods to build training dataset. The proposed method combines the strengths of different SR methods and generates better results.  
(3) The proposed SR framework is highly flexible and produce diverse results given different rank datasets, perceptual metrics, and loss combinations.

\section{Related work}
\label{sec:2}
\textbf{Super resolution.} Since Dong et al. \cite{dong2014learning} first introduced convolutional neural networks (CNNs) to the SR task, a series of learning-based works \cite{zhang2012single,haris2018deep,kim2016accurate, He_2019_CVPR, feng2019suppressing, gu2019blind} have achieved great improvements in terms of PSNR. For example, Kim et al. \cite{kim2016accurate} propose a deep network VDSR with gradient clipping. The residual and dense block  \cite{lim2017enhanced,zhang2018residual} are explored to improve the super-resolved results. In addition, SRGAN \cite{ledig2017photo} is proposed to generate more realistic images. Then, texture matching  \cite{sajjadi2017enhancenet} and semantic prior \cite{wang2018recovering} are introduced to improve perceptual quality. Furthermore, the perceptual index  \cite{blau20182018} consisting of NIQE \cite{mittal2013making} and  Ma \cite{ma2017learning} is adopted to measure the perceptual SR methods in the PIRM2018-SR Challenge at ECCV \cite{blau20182018}. In the Challenge, ESRGAN \cite{Wang_2018_ECCV_Workshops} achieves the state-of-the-art performance by improving network architecture and loss functions. 

\textbf{CNN for NR-IQA.} No-reference Image Quality Assessment (NR-IQA) can be implemented by learning-based models, which extract hand-crafted features from Natural Scene Statistics (NSS), such as CBIQ \cite{ye2011no}, NIQE \cite{mittal2013making}, and Ma \cite{ma2017learning}, etc. In  \cite{li2011blind}, Li et al. develop a general regression neural network to fit human subjective opinion scores with pre-extracted features. Kang et al.  \cite{kang2014convolutional,bosse2018deep} integrate a general CNN framework which can predict image quality on local regions. In addition, Liu et al.  \cite{liu2017rankiqa} propose RankIQA to tackle the problem of lacking human-annotated data in NR-IQA. They first generate large distorted images in different distortion level. Then they train a Siamese Network to learn the rank of the quality of those images, which can improve the performance of the image quality scores.

\textbf{Learning to rank.} It has been demonstrated that learning to rank approach is effective in computer vision. For instance, Devi Parikh et al.  \cite{parikh2011relative} model relative attributes using a well-learned ranking function. Yang et al. \cite{yang2016deep} first employ CNN for relative attribute ranking in a unified framework. One of the most relevant studies to our work is RankCGAN \cite{saquil2018ranking}, which investigates the use of GAN to tackle the task of image generation with semantic attributes. Unlike standard GANs that generate the image from noise input (CGAN \cite{mirza2014conditional}), RankCGAN incorporates a pairwise Ranker into CGAN architecture so that it can handle continuous attribute values with subjective measures. 

\begin{figure*}[h!]
\setlength{\abovecaptionskip}{-0.4cm}
\setlength{\belowcaptionskip}{-0.2cm}
\begin{center}
\includegraphics[width=1\linewidth]{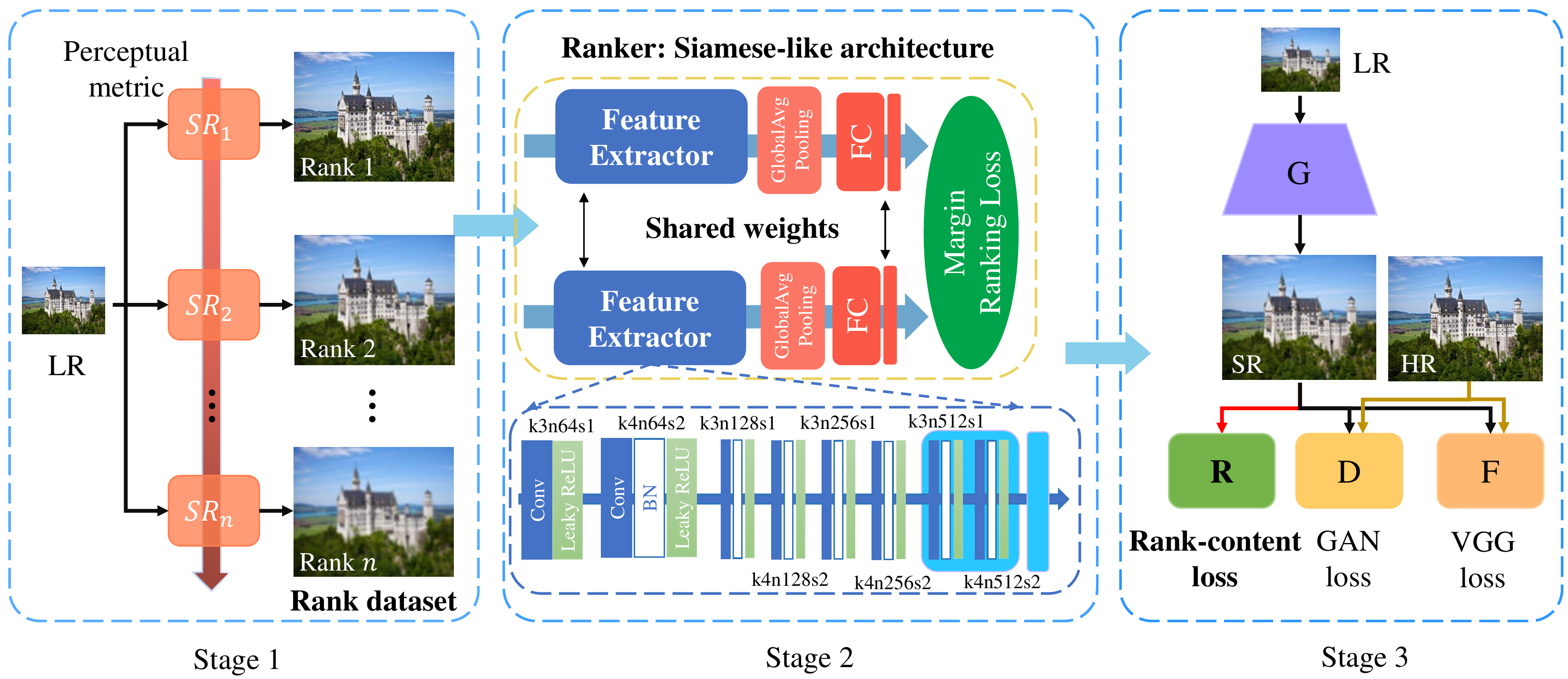} 
\end{center}
   \caption{\textbf{Overview of the proposed method.} \textbf{Stage 1:} Generate pair-wise rank images by different SR models in the orientation of perceptual metrics. \textbf{Stage 2:} Train Siamese-like Ranker network. \textbf{Stage 3:} Introduce rank-content loss derived from well-trained Ranker to guide GAN training. RankSRGAN consists of a generator(G), discriminator(D), a fixed Feature extractor(F) and Ranker(R). }
\label{fig:2}
\vskip -0.25cm
\end{figure*}

\section{Method}
\label{sec:3}

\subsection{Overview of RankSRGAN}
The proposed framework is built upon the GAN-based \cite{ledig2017photo} SR approach, which consists of a generator and a discriminator. The discriminator network tries to distinguish the ground-truth images from the super-resolved results, while the generator network is trained to fool the discriminator. To obtain more natural textures, we propose to add additional constraints on the standard SRGAN \cite{ledig2017photo} by exploiting the prior knowledge of perceptual metrics to improve the visual quality of output images. The overall framework of our approach is depicted in Figure \ref{fig:2}. The pipeline involves the following three stages:

\textbf{Stage 1: Generate pair-wise rank images.} First, we employ different SR methods to generate super-resolved images on public SR datasets. Then we apply a chosen perceptual metric (e.g. NIQE) on the generated images. After that, we can pick up two images of the same content to form a pair and rank the pair-wise images according to the quality score calculated by the perceptual metric. Finally, we obtain the pair-wise images and the associated ranking labels. More details will be presented in Section \ref{4.1}.

\textbf{Stage 2: Train Ranker.} The Ranker adopts a Siamese architecture to learn the beheviour of perceptual metrics and the network structure is depicted in Section \ref{section:3.2}. We adopt margin-ranking loss, which is commonly used in ``learning to rank'' \cite{burges2005learning}, as the cost function to optimize Ranker. The learned Ranker is supposed to have the ability to rank images according to their perceptual scores.

\textbf{Stage 3: Introduce rank-content loss.} Once the Ranker is well-trained, we use it to define a rank-content loss for a standard SRGAN to generate visually pleasing images. Please see the rank-content loss in Section \ref{3.3}.

\subsection{Ranker}
\label{section:3.2}
\textbf{Rank dataset.} Similar to \cite{choi2018deep,liu2017rankiqa}, we use super-resolution results of different SR methods to represent different perceptual levels.
With a given perceptual metric, we can rank these results in a pair-wise manner. Picking any two SR images, we can get their ranking order according to the quality score measured by the perpetual metric. These pair-wise data with ranking labels form a new dataset, which is defined as the rank dataset. Then we let the proposed Ranker learn the ranking orders. Specifically, given two input images $y_{1}$, $y_{2}$, the ranking scores $s_1$ and $s_2$ can be obtained by

\begin{small}
\begin{equation}
{s_1} = R({y_1};{\Theta _R})
\end{equation}
\begin{equation}
{s_2} = R({y_2};{\Theta _R}),
\end{equation}
\end{small}where ${\Theta _R}$ represents the network weights and $R(.)$ indicates the mapping function of Ranker. In order to make the Ranker output similar ranking orders as the perceptual metric, we can formulate:
\begin{small}
\begin{equation}
\left\{
             \begin{array}{lr}
             s_1<s_2  & if \quad m_{y_1}<m_{y_2}\\
             s_1>s_2  & if \quad m_{y_1}>m_{y_2}\\ 
             \end{array}
\right.,
\end{equation}
\end{small}where $ m_{y_1} $ and $ m_{y_2} $ represent the quality scores of image $y_1$ and image $y_2$, respectively. A well-trained Ranker could guide the SR model to be optimized in the orientation of the given perceptual metric.

\textbf{Siamese architecture.} The Ranker uses a Siamese-like architecture \cite{bromley1994signature,chopra2005learning,zagoruyko2015learning}, which is effective for pair-wise inputs. The architecture of Ranker is shown in Figure \ref{fig:2}. It has two identical network branches which contain a series of convolutional, LeakyReLU, pooling and full-connected layers. Here we use a Global Average Pooling layer after the Feature Extractor, thus the architecture can get rid of the limit of input size. To obtain the ranking scores, we employ a fully-connected layer as a regressor to quantify the rank results. Note that we do not aim to predict the real values of the perceptual metric since we only care about the ranking information.  Finally, the outputs of two branches are passed to the margin-ranking loss module, where we can compute the gradients and apply back-propagation to update parameters of the whole network.

\textbf{Optimization.} To train Ranker, we employ margin-ranking loss that is commonly used in sorting problems \cite{yang2016deep,liu2017rankiqa}. The margin-ranking loss is given below:
\begin{small}
\begin{equation}
\setlength{\abovedisplayskip}{3pt} 
\setlength{\belowdisplayskip}{3pt}
\begin{split}
L({s_1},&{s_2};\gamma) = \max (0,({s_1} - {s_2})*\gamma + \varepsilon )\\
&\left\{
             \begin{array}{lr}
             \gamma\,\,=-1  & if \quad m_{y_1}<m_{y_2}\\
             \gamma\,\,=1  & if \quad m_{y_1}>m_{y_2}\\ 
             \end{array}
\right.,
\end{split}
\end{equation}
\end{small}where the $s_1$ and $s_2$ represent the ranking scores of pair-wise images. The $\gamma$ is the rank label of the pair-wise training images. The margin $\varepsilon$ can control the distance between $s_1$ and $s_2$. Therefore, the $N$ pair-wise training images can be optimized by:

\begin{small}
\begin{equation}
\setlength{\abovedisplayskip}{-9pt} 
\setlength{\belowdisplayskip}{-5pt}
\begin{split}
\hat{\Theta}=&\mathop{\arg\min}_{\Theta_R}\frac{1}{N}\sum_{i=1}^N L(s_1^{(i)},s_2^{(i)};\gamma^{(i)})\\
=&\mathop{\arg\min}_{\Theta_R}\frac{1}{N}\sum_{i=1}^NL(R({y_1^{(i)}};{\Theta _R}),R({y_2^{(i)}};{\Theta _R});\gamma^{(i)})
\end{split}
\end{equation}
\end{small}

\subsection{RankSRGAN}
\label{3.3}
RankSRGAN consists of a standard SRGAN and the proposed Ranker, as shown in Figure \ref{fig:2}. Compared with existing SRGAN, our framework simply adds a well-trained Ranker to constrain the generator in SR space. To obtain visually pleasing super-resolved results, adversarial learning \cite{ledig2017photo,sajjadi2017enhancenet} is applied to our framework where the generator and discriminator are jointly optimized with the objective given below:
\begin{small}
\begin{equation}
\setlength{\abovedisplayskip}{3pt} 
\setlength{\belowdisplayskip}{3pt}
\mathop{\min}_{\theta}\mathop{\max}_{\eta} E_{y\backsim p_{HR}}log{D_{\eta}}(y)+E_{y\backsim p_{LR}}log(1-D_\eta(G_\theta(x))),
\end{equation}
\end{small}where $ p_{HR} $ and $ p_{LR} $ represent the probability distributions of $ HR $ and $ LR $ samples, respectively. In order to demonstrate the effectiveness of the proposed Ranker, we do not use complex architectural designs of GAN \cite{Wang_2018_ECCV_Workshops} but use the general SRGAN \cite{ledig2017photo}.

\textbf{Perceptual loss.} In \cite{dosovitskiy2016generating,johnson2016perceptual}, the perceptual loss is proposed to measure the perceptual similarity between two images. Instead of computing distances in image pixel space, the images are first mapped into feature space and the perceptual loss can be presented as:
\begin{small}
\begin{equation}
\setlength{\abovedisplayskip}{3pt} 
\setlength{\belowdisplayskip}{1pt}
L_P=\sum_i||\phi(\hat{y_i})-\phi(y_i)||_2^2,
\end{equation}
\end{small}where $\phi(y_i)$ and $\phi(\hat{y_i})$ represent the feature maps of HR and SR images, respectively. Here $\phi$ is obtained by the 5-th convolution (before maxpooling) layer within VGG19 network \cite{simonyan2014very}.

\textbf{Adversarial loss.} Adversarial training \cite{ledig2017photo,sajjadi2017enhancenet} is recently used to produce natural-looking images. A discriminator is trained to distinguish the real image from the generated image. This is a minimax game approach where the generator loss $ L_G $ is defined based on the output of discriminator:
\begin{small}
\begin{equation}
\setlength{\abovedisplayskip}{4pt} 
\setlength{\belowdisplayskip}{4pt}
L_G=-logD(G(x_i)),
\end{equation}
\end{small}where $ x_i $ is the $ LR $ image, $ D(G(x_i)) $ represents the probability of the discriminator over all training samples.

\textbf{Rank-content loss.} The generated image is put into the Ranker to predict the ranking score. Then, the rank-content loss can  be defined as:
\begin{equation}
\setlength{\abovedisplayskip}{4pt} 
\setlength{\belowdisplayskip}{4pt}
L_R=sigmoid(R(G(x_i))),
\end{equation}
where $ R(G(x_i)) $ is the ranking score of the generated image. A lower ranking score indicates better perceptual quality. After applying the sigmoid function, $ L_R $ represents ranking-content loss ranging from $0$ to $1$.

\subsection{Analysis of Ranker}
\label{3.4}
The proposed Ranker possesses an appealing property: by elaborately selecting the SR algorithms and the perceptual metric, the RankSRGAN has the potential to surpass the upper bound of these methods and achieve superior performance. To validate this comment, we select the state-of-the-art perceptual SR methods -- SRGAN \cite{ledig2017photo} and ESRGAN \cite{Wang_2018_ECCV_Workshops} to build the rank dataset.  Then we use the perceptual metric NIQE \cite{mittal2013making} for evaluation. NIQE is demonstrated to be highly correlated with human ratings and easy to implement. A lower NIQE value indicates better perceptual quality. When measured with NIQE on the PIRM-Test \cite{blau20182018} dataset, the average scores of SRGAN and ESRGAN are 2.70 and 2.55, respectively. ESRGAN obtains better NIQE scores for most images but not all images, indicating that SRGAN and ESRGAN have mixed ranking orders with NIQE.

\begin{figure}[t]
\setlength{\abovecaptionskip}{-0.2cm}
\setlength{\belowcaptionskip}{-0.2cm}
\begin{center}
	\includegraphics[width=1\linewidth]{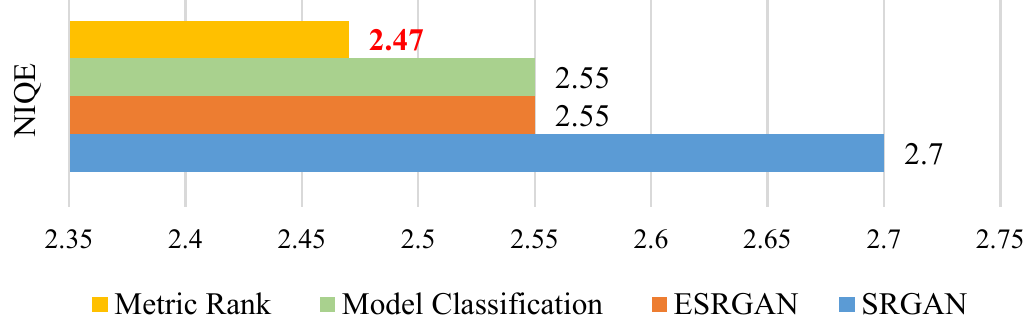} 

\end{center}
   \caption{The upper bound (average NIQE value) of SRGAN, ESRGAN, model rank and model classification.}
\label{fig:4}
\vskip -0.4cm
\end{figure}

In order to examine the effectiveness of our proposed Ranker, we compare two ranking strategies -- metric rank and model classification. Metric rank, which is our proposed method, uses perceptual metrics to rank the images. For example, in each image pair, the one with a lower score is labeled to 1 and the other is 2. The model classification, as the comparison method, ranks images according to the used SR methods, i.e., all results of ESRGAN are labeled to 1 and those of SRGAN are labeled to 2. We then give an analysis of the upper bound of these two methods. The upper bound can be calculated as:
\begin{small}
\begin{equation}
\setlength{\abovedisplayskip}{5pt}
\setlength{\belowdisplayskip}{5pt}
\begin{split}
UB_{MC}= Mean(PM_{SR2-L} + PM_{SR2-H} )\\
UB_{MR}= Mean(PM_{SR2-L} + PM_{SR1-L})\\
where: PM_{SR1-L} < PM_{SR2-H},
\end{split}
\label{equation:10}
\end{equation}
\end{small}where $UB_{MC}$ and $UB_{MR}$ represent the upper bound of model classification and metric rank, respectively. $PM$ (Perceptual Metric) is the perceptual score for each image in the corresponding class. (SR1, SR2) represents two SR results of the same LR. Subscripts $-L$ and $-H$ indicate the Lower and Higher perceptual score in (SR1, SR2). We use (SRGAN, ESRGAN) as (SR1, SR2) to obtain the upper bound of these methods, as shown in Figure \ref{fig:4}. Obviously, metric rank could combine the better parts of different algorithms and exceed the upper bound of a single algorithm.

We further conduct SR experiments to support the above analysis. We use the metric rank and model classification approach to label the rank dataset. Then the Ranker-MC (model classification) and Ranker-MR (metric rank) are used to train separate RankSRGAN models. Figure \ref{fig:5} shows the quantitative results (NIQE), where RankSRGAN-MR outperforms ESRGAN and RankSRGAN-MC. This demonstrates that our method can exceed the upper bound of all chosen SR algorithms.

\section{Experiments}
\subsection{Training details of Ranker}
\label{4.1}
\textbf{Datasets.} We use DIV2K (800 images) \cite{agustsson2017ntire} and Flickr2K (2650 images) \cite{agustsson2017ntire} datasets to generate pair-wise images as rank dataset for training. Three different SR algorithms (SRResNet \cite{ledig2017photo}, SRGAN \cite{ledig2017photo} and ESRGAN \cite{Wang_2018_ECCV_Workshops}) are used to generate super-resolved images as three perceptual levels, as shown in Table \ref{table:1}. 
\vskip -0.2cm

\begin{table}[!ht]
\small 
\setlength{\abovecaptionskip}{-3pt}
\setlength{\belowcaptionskip}{-10pt}
\begin{center}
\begin{tabular}{c|ccc}
\hline\hline
PIRM-Test & SRResNet & SRGAN & ESRGAN \\
\hline
NIQE & 5.968 & 2.705 & 2.557\\
PSNR & 28.33 & 25.62 & 25.30\\
\hline
\end{tabular}
\end{center}
\caption{The performance of super-resolved results with three perceptual levels in PIRM-Test \cite{blau20182018}}
\label{table:1}
\vskip -0.1cm
\end{table}
We extract patches from those pair-wise images with a stride of $ 200 $ and size of $ 296\times 296 $. For one perceptual level (SR algorithm), we can generate 150 K patches (10$\%$ for validation, 90$\%$ for training). Inspired by PIRM2018-SR Challenge \cite{blau20182018}, we use NIQE \cite{mittal2013making} as the perceptual metric, while other metrics will be investigated in Section \ref{section4.5}. Finally, we label every image pair to (3,2,1) according to the order of corresponding NIQE value (the one with the best NIQE value is set to 1).
\begin{figure}[t]
\setlength{\abovecaptionskip}{-0.1cm}
\setlength{\belowcaptionskip}{-0.6cm}
\begin{center}
	\includegraphics[width=1\linewidth]{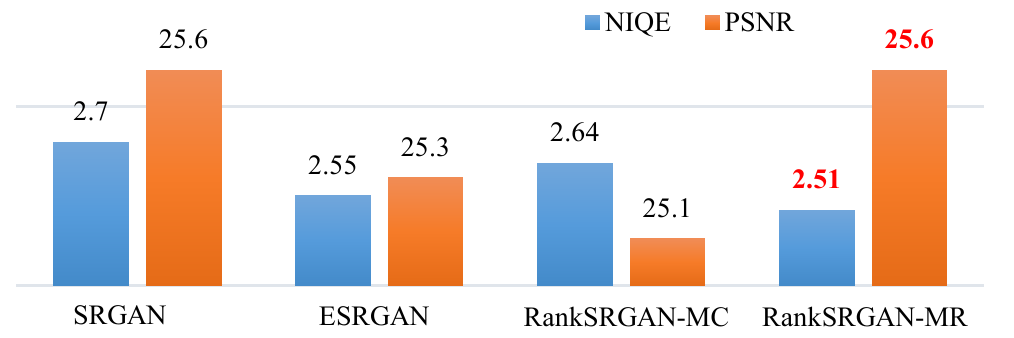} 

\end{center}
   \caption{The NIQE of RankSRGAN-MR exceeds that of SRGAN, ESRGAN and RankSRGAN-MC.}
\label{fig:5}
\end{figure}

\textbf{Implementation details.} As shown in Figure \ref{fig:2}, we  utilize VGG \cite{simonyan2014very} structure to implement the Ranker \cite{liu2017rankiqa}, which includes 10 convolutional layers, a series of batch normalization and LeakyReLU operations. Instead of max-pooling, we apply convolutional layer with a kernel size $4$ and stride $2$ to downsample the features. In one iteration, two patches with different perceptual levels are randomly selected as the input of Ranker. 
For optimization, we use Adam \cite{kingma2014adam} optimizer with weight decay $1\times10^{-4}$. The learning rate is initialized to $1\times10^{-3}$ and decreases with a factor $0.5$ of every $10\times10^{4}$ iterations for total $30\times10^{4}$ iterations. The margin $\epsilon$ of margin-ranking loss is set to $0.5$. For weight initialization, we use He. \cite{he2015delving} method to initialize the weights of Ranker. 

\textbf{Evaluation.} The Spearman Rank Order Correlation Coefficient (SROCC) \cite{liu2017rankiqa} is a traditional evaluation metric to evaluate the performance of image quality assessment algorithms. In our experiment, SROCC is employed to measure the monotonic relationship between the label and the ranking score. Given $N$ images, the SROCC is computed as:
\begin{small}
\begin{equation}
SROCC=1 - \frac{6\sum_{i=1}^{N}(y_i-\hat{y_i})^2}{N(N^2-1)},
\end{equation}
\end{small}where $y_i$ represents the order of label, and $\hat{y_i}$ is the order of output score of the Ranker. SROCC has the ability to measure the accuracy of Ranker. The larger value of SROCC represents the better accuracy of Ranker. For validation dataset, the ranker achieves a SROCC of ${0.88}$, which is an adequate performance compared with those in the related work \cite{choi2018deep, liu2017rankiqa}.

\begin{table*}
\setlength{\abovecaptionskip}{0.2cm}
\setlength{\belowcaptionskip}{-0.8cm}
\begin{center}
\begin{tabular}{c|c|cccccc}
\hline\hline
Dataset & Metric & Bicubic & FSRCNN & SRResNet & SRGAN & ESRGAN & RankSRGAN (ours)\\
\hline
\multirow{3}*{Set14}&NIQE & 7.61 &6.92&6.12&3.82&3.28&\textbf{3.28}\\
       & PI & 6.97 & 6.16 &5.36&2.98&2.61&\textbf{2.61}\\
			   & PSNR & 26.08 & 27.66 &\textbf{28.57}&26.68&26.39&26.57\\
\hline
\multirow{3}*{BSD100}&NIQE & 7.60 &7.11&6.43&3.29&3.21&\textbf{3.01}\\
        & PI & 6.94 & 6.17 &5.34&2.37&2.27&\textbf{2.15}\\
	            & PSNR & 25.96 & 26.94 &\textbf{27.61}&25.67 &25.72 &25.57\\
\hline
\multirow{3}*{PIRM-Test}&NIQE & 7.45 &6.86& 5.98&2.71&2.56& \textbf{2.51}  \\
           & PI & 7.33 & 6.02 &5.18&2.09&1.98&\textbf{1.95}\\
          		   & PSNR & 26.45 & 27.57 &\textbf{28.33}&25.60&25.30&25.62\\
\hline
\end{tabular}

\caption{\label{table:2}Average NIQE \cite{mittal2013making}, PI \cite{blau20182018} and PSNR values on the Set14 \cite{zeyde2010single}, BSD100 \cite{martin2001database} and PIRM-Test \cite{blau20182018}.}

\end{center}

\end{table*}


\begin{figure*}[h]
\setlength{\abovecaptionskip}{-0.2cm}
\setlength{\belowcaptionskip}{-0.1cm}
\begin{center}
\includegraphics[width=1\linewidth]{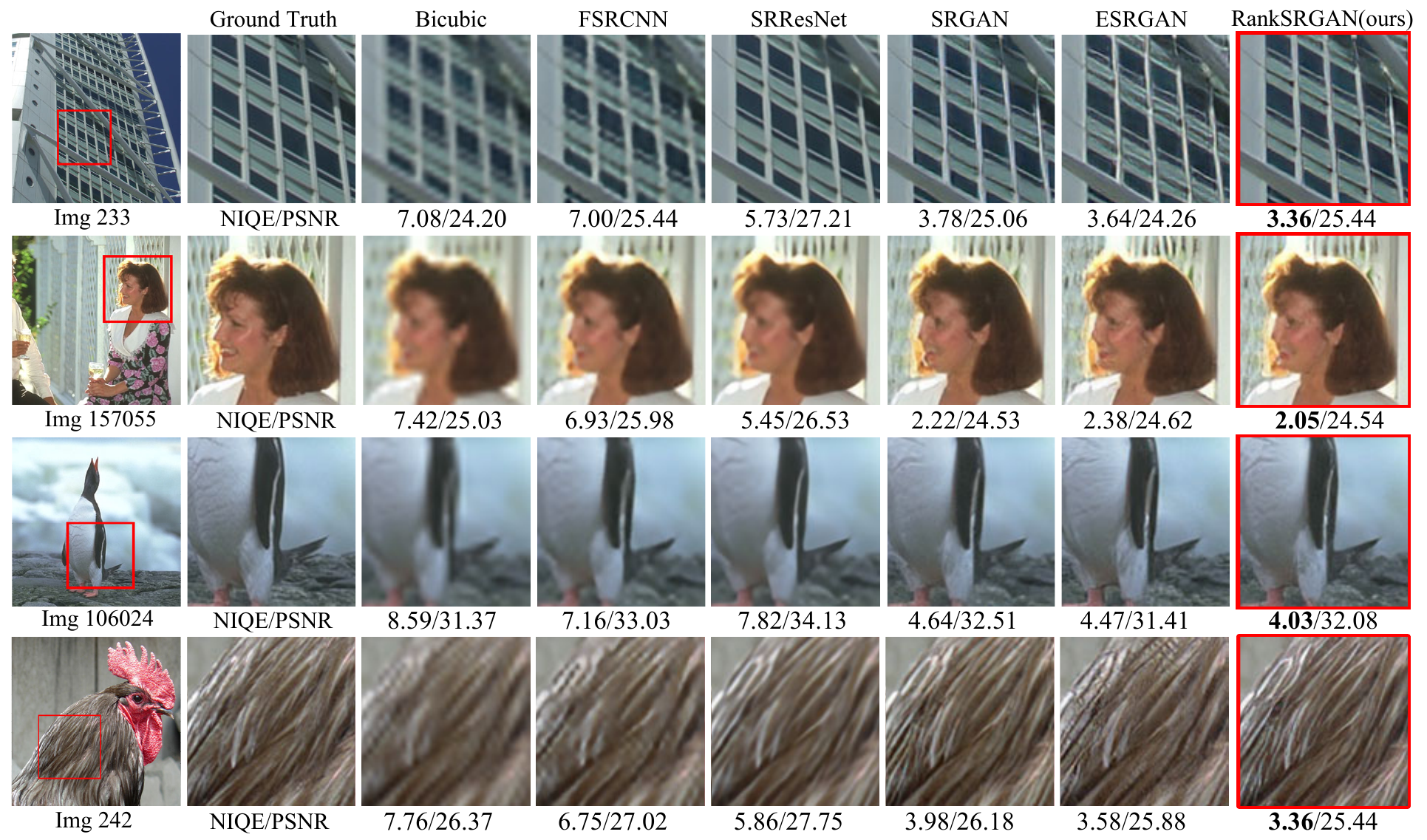} 
\end{center}
   \caption{Visual results comparison of our model with other works on $\times 4$ super-resolution. Lower NIQE value indicates better perceptual quality, while higher PSNR indicates less distortion.}
\label{fig:Visual results}
\vskip -0.5cm
\end{figure*}

\subsection{Training details of RankSRGAN}
\label{4.2}
We use the DIV2K \cite{agustsson2017ntire} dataset to train RankSRGAN. The patch sizes of HR and LR are set to 296 and 74, respectively. For testing, we use benchmark datasets Set14 \cite{zeyde2010single}, BSD100 \cite{martin2001database} and PIRM-test \cite{blau20182018}. PIRM-test is used to measure the
perceptual quality of SR methods in PIRM2018-SR \cite{blau20182018}. Following the settings of SRGAN \cite{ledig2017photo}, we employ a standard SRGAN \cite{ledig2017photo} as our base model. The generator is built with 16 residual blocks, and the batch-normalization layers are removed \cite{Wang_2018_ECCV_Workshops}. The discriminator utilizes the VGG network \cite{simonyan2014very} with ten convolutional layers. The mini-batch size is set to 8. At each training step, the combination of loss functions (Section \ref{3.3}) for the generator is:
\vskip -0.2cm
\begin{equation}
\setlength{\abovecaptionskip}{-0.4cm} 
\setlength{\belowcaptionskip}{-0.4cm}
L_{total}=L_P+0.005L_G+0.03L_R,
\end{equation}
where the weights of $L_G$ and $L_R$ are determined empirically to obtain high perceptual improvement\cite{choi2018deep,ledig2017photo,Wang_2018_ECCV_Workshops}. The Adam\cite{kingma2014adam} optimization method with $ \beta_1=0.9 $ is used for training. For generator and discriminator, the initial learning rate is set to $1 \times 10^{-4} $ which is reduced by a half for multi-step $ [50\times10^3,100\times10^3,200\times10^3,300\times10^3]$. A total of $ 600\times10^3$ iterations are executed by PyTorch. 
In training procedure, we add Ranker to the standard SRGAN. The Ranker takes some time to predict the ranking score, thus the traning time is a little slower (about 1.18 times) than standard SRGAN\cite{ledig2017photo}. For the generator, the number of parameters remains the same as SRGAN\cite{ledig2017photo}.

\subsection{Comparison with the-state-of-the-arts}
\label{section4.3}
We compare the performance of the proposed method with the state-of-the-art perceptual SR methods ESRGAN \cite{Wang_2018_ECCV_Workshops}/ SRGAN \cite{ledig2017photo} and the PSNR-orientated methods FSRCNN \cite{dong2016accelerating} and SRResNet \cite{ledig2017photo} \footnote{Our implementation of SRResNet and SRGAN achieve even better performance than that reported in the original paper.}. The evaluation metrics include NIQE \cite{mittal2013making}, PI \cite{blau20182018} and PSNR. Table \ref{table:2} shows their performance on three test datasets -- Set14, BSD100 and PIRM-Test. Note that lower NIQE/PI indicates better visual quality. When comparing our method with SRGAN and ESRGAN, we find that RankSRGAN achieves the best NIQE and PI performance on all test sets. Furthermore, the improvement of perceptual scores does not come at the price of PSNR. Note that in PIRM-Test, RankSRGAN also obtains the highest PSNR values among perceptual SR methods. Figure \ref{fig:Visual results} shows some visual examples, where we observe that our method could generate more realistic textures without introducing additional artifacts (please see the windows in Img 233 and feathers in Img 242). 

\begin{figure}[t]
\setlength{\abovecaptionskip}{-0.4cm}
\setlength{\belowcaptionskip}{-0.5cm}
\begin{center}
	\includegraphics[width=1\linewidth]{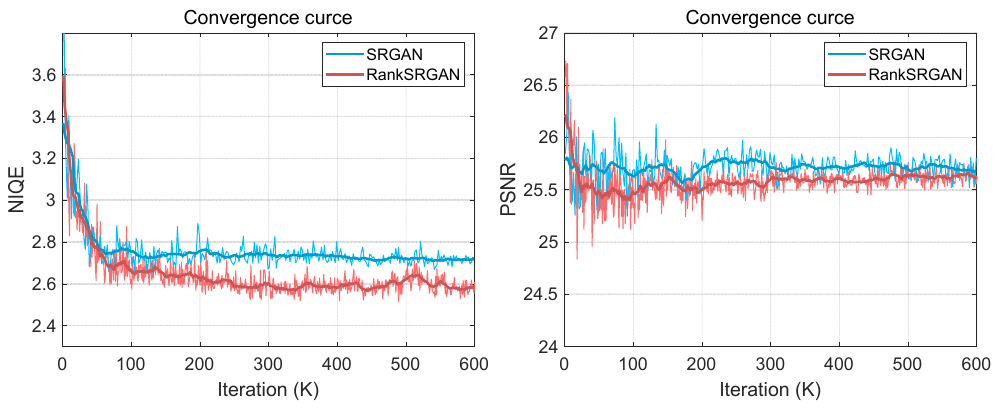} 

\end{center}
   \caption{Convergence curves of RankSRGAN in PSNR/ NIQE.}
\label{fig:curve}
\end{figure}

As the results may vary across different iterations, we further show the convergence curves of RankSRGAN in Figure \ref{fig:curve}. Their performance on NIQE and PSNR are relatively stable during the training process. For PSNR, they obtain comparable results. But for NIQE, RankSRGAN is consistently better than SRGAN by a large margin. 

\subsection{Ablation study}
\label{section4.5}
\textbf{Effect of different rank datasets.} The key factor that influences the performance of Ranker is the choice of SR algorithms. In the main experiments, we use (SRResNet, SRGAN, ESRGAN) to generate the rank dataset. Then what if we select other SR algorithms? Will we always obtain better results than SRGAN? To answer the question, we first analyze the reason of using these three algorithms, then conduct another experiment using a different combination. 

As our baseline model is SRGAN, we need the Ranker to have the ability to rank the outputs of SRGAN. Since the training of SRGAN starts from the pre-trained model SRResNet, the Ranker should recognize the results between SRResNet and SRGAN. That is the reason why we choose SRResNet and SRGAN. Then the next step is to find a better algorithm that could guide the model to achieve better results. We choose ESRGAN as it surpasses SRGAN by a large margin in the PIRM-SR2018 challenge \cite{blau20182018}. Therefore, we believe that a better algorithm than SRGAN could always lead to better performance. 
\vskip -0.1cm

\begin{table}[h]
\renewcommand\tabcolsep{2.0pt}
\small
\setlength{\abovecaptionskip}{-0.2cm}
\setlength{\belowcaptionskip}{-0.3cm}
\begin{center}
\begin{tabular}{l|ccc}
\hline\hline
Method &SRGAN&RankSRGAN&RankSRGAN-HR \\
\hline
NIQE&2.70&\textbf{2.51}&2.58\\
PSNR&25.62&25.60&\textbf{26.00}\\
\hline
\end{tabular}
\end{center}
\caption{Comparison with RankSRGAN and RankSRGAN-HR.}
\label{table:RankSRGAN-HR}
\end{table}

To validate this comment, we directly use the ground truth HR as the third algorithm, which is the extreme case. We still apply NIQE for evaluation. Interestingly, although HR images have infinite PSNR values, they cannot surpass all the results of SRGAN on NIQE. Similar to ESRGAN, HR and SRGAN have mixed ranking orders. We train our Ranker with (SRResNet, SRGAN, HR) and obtain the new SR model -- RankSRGAN-HR. Table \ref{table:RankSRGAN-HR} compares its results with SRGAN and RankSRGAN. As expected, RankSRGAN-HR achieves better NIQE values than SRGAN. But at the same time, RankSRGAN-HR also improves the PSNR by almost 0.4 dB. It achieves a good balance between the perceptual metric and PSNR. This also indicates that the model could always have an improvement space as long as we have better algorithms for guidance.

\textbf{Effect of different perceptual metrics.} As we claim that Ranker can guide the SR model to be optimized in the direction of perceptual metrics, we need to verify whether it works for other perceptual metrics. We choose Ma \cite{martin2001database} and PI \cite{blau20182018}, which show high correlation with Mean-Opinion-Score (Ma: 0.61, PI: 0.83) in \cite{blau20182018}. We use Ma and PI as the evaluation metric to generate the rank dataset. All other settings remain the same as RankSRGAN with NIQE. The only difference in these experiments is the ranking labels in the rank dataset. The results are summarized in Table \ref{table:RankSRGAN-metric}, where we observe that the Ranker could help RankSRGAN achieve the best performance in the chosen metric. This shows that our method can generalize well on different perceptual metrics.

\begin{table}[h]
\begin{center}
\small 
\setlength{\abovecaptionskip}{-0.3cm}
\setlength{\belowcaptionskip}{-0.4cm}
\begin{tabular}{l|cccc}
\hline\hline
Method &NIQE&10-Ma&PI&PSNR \\
\hline
SRGAN&2.71&1.47&2.09&25.62\\
ESRGAN&2.56&1.40&1.98&25.30\\
\hline
RankSRGAN$_N$&\textbf{2.51}&1.39&1.95&25.62\\
RankSRGAN$_M$&2.65&\textbf{1.38}&2.01&25.21\\
RankSRGAN$_{PI}$&2.49&1.39&\textbf{1.94}&25.49\\
\hline
\end{tabular}
\end{center}
\caption{The performance of RankSRGAN with different Rankers. $N$: Ranker with NIQE \cite{mittal2013making}, $M$: Ranker with Ma \cite{ma2017learning} and $PI$: Ranker with PI \cite{blau20182018}.}
\label{table:RankSRGAN-metric}
\end{table}
\vskip -0.3cm

\textbf{Effect of Ranker: Rank VS. Regression.} To train our Ranker, we choose to use the ranking orders instead of the real values of the perceptual metric. Actually, we can also let the network directly learn the real values. In \cite{choi2018deep}, Choi et al. use a regression network to predict a subjective score for a given image and define a corresponding subjective score loss. To compare these two strategies, we train a ``regression'' Ranker with MSE loss instead of the margin-ranking loss. The labels in the rank dataset are real values of the perceptual metric. We use NIQE and Ma to generate the labels of rank dataset. All the other settings remain the same as RankSRGAN. 
\begin{table}[h]
\small 
\setlength{\abovecaptionskip}{-0.2cm} 
\setlength{\belowcaptionskip}{-0.4cm}
\begin{center}
\begin{tabular}{l|c c}
\hline\hline
Metric&Method&$E(|SR_1-SR_2|)$\\
\hline
NIQE&regression&0.06\\
NIQE&rank&\textbf{0.11}\\
\hline
Ma&regression&0.09\\
Ma&rank&\textbf{0.15}\\
\hline
\end{tabular}
\end{center}
\caption{The distance between $SR_1$ and $SR_2$ with regression and rank.}
\label{table:Ranker-distance}
\end{table}

Theoretically, the real values of perceptual metrics may distribute unevenly among different algorithms. For example, SRGAN and ESRGAN are very close to each other on NIQE values. This presents a difficulty for the learning of regression. On the contrary, learning ranking orders can simply ignore these variances. In experiments, we first measure the distances between the outputs of SRGAN and ESRGAN with different strategies. Table \ref{table:Ranker-distance} shows the mean absolute distances of these two strategies. Obviously, results with rank have larger distances than results with regression. When applying these Rankers in SR training, the rank strategy achieves better performance than the regression strategy on the selected perceptual metric. Results are shown in Table \ref{table:RankSRGAN-regression}.
\vskip -0.1cm
\begin{table}[h]
\small 
\setlength{\abovecaptionskip}{-0.1cm}
\setlength{\belowcaptionskip}{-0.3cm}
\begin{center}
\begin{tabular}{l|ccc}
\hline\hline
Method&NIQE&10-Ma&PSNR \\
\hline
SRGAN&2.71&1.47&25.62\\
ESRGAN&2.55&1.40&25.30\\
\hline
RankSRGAN-Re$_N$&2.53&1.42&25.58\\
RankSRGAN$_N$&\textbf{2.51}&1.39&25.60\\
\hline
RankSRGAN-Re$_M$&2.61&1.43&25.23\\
RankSRGAN$_M$&2.65&\textbf{1.38}&25.21\\
\hline
\end{tabular}
\end{center}
\caption{The performance of RankSRGAN with different rankers. Re: Ranker with regression, $N$: Ranker with NIQE, $M$: Ranker with Ma.}
\label{table:RankSRGAN-regression}
\end{table}

\textbf{Effect of different losses.} To test the effects of rank-content loss, we expect to add MSE loss to achieve improvement in PSNR. Table \ref{table:different loss} shows the performance of our method trained with the combination of loss functions. 

\begin{table}[h]
\small 
\setlength{\abovecaptionskip}{-0.1cm} 
\setlength{\belowcaptionskip}{-0.4cm}
\begin{center}
\begin{tabular}{l|c|cc}
\hline\hline
Method & Loss &NIQE&PSNR \\
\hline
SRGAN&$L_P$&2.71&25.62\\
ESRGAN&$L_P+10L_M$&2.55&25.30\\

\hline
RankSRGAN&$L_P+L_{R}$&\textbf{2.51}&25.62\\
RankSRGAN-M$_1$&$L_P+L_{R}+\alpha_1L_M$&2.55&25.87\\
RankSRGAN-M$_2$&$L_P+L_{R}+\alpha_2L_M$&2.72&\textbf{26.62}\\

\hline
\end{tabular}
\end{center}
\caption{The performance of RankSRGAN with the combination of loss functions (P: perceptual loss, R: rank-content loss, M: MSE loss). $\alpha_1, \alpha_2 : \left\{1,5\right\}$ }
\label{table:different loss}
\end{table}
As expected, increasing the contribution of the MSE loss with the larger $\alpha$ results in higher PSNR values. On the other hand, the NIQE values are increased which is a tradeoff between PSNR and NIQE as mentioned in \cite{blau2018perception}, and our method has a capability to deal with the priorities by adjusting the weights of the loss functions.

\subsection{User study}
\label{section: user study}

To demonstrate the effectiveness and superiority of RankSRGAN, we conduct a user study against state-of-the-art models, i.e. SRGAN \cite{ledig2017photo} and ESRGAN \cite{Wang_2018_ECCV_Workshops}. In the first session, two different SR images are shown at the same time where one is generated by the proposed RankSRGAN and the other is generated by SRGAN or ESRGAN. The participants are required to pick the image that is more visually pleasant (more natural and realistic). We use the PIRM-Test \cite{blau20182018} dataset as the testing dataset. There are a total of 100 images, from which 30 images are randomly selected for each participant. To make a better comparison, one small patch from the image is zoomed in. 
In the second session, we focus on the perceptual quality of different typical SR methods in a sorting manner. The participants are asked to rank 4 versions of each image: SRResNet \cite{ledig2017photo}, ESRGAN \cite{Wang_2018_ECCV_Workshops}, RankSRGAN, and the Ground Truth (GT) image according to their visual qualities. Similar to the first session, 20 images are randomly shown for each participant. There are totally 30 participants to finish the user study. 
\begin{figure}
\setlength{\abovecaptionskip}{-0.1cm}
\setlength{\belowcaptionskip}{-0.5cm}
\begin{center}
\includegraphics[width=1\linewidth]{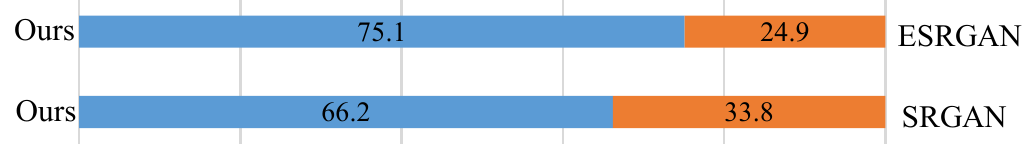} 
\end{center}
   \caption{The results of user studies, comparing our method with SRGAN  \cite{ledig2017photo} and ESRGAN \cite{Wang_2018_ECCV_Workshops}.}
\label{fig:user study1}
\end{figure}

\begin{figure}[h]
\setlength{\abovecaptionskip}{-0.1cm}
\setlength{\belowcaptionskip}{-0.3cm}
\begin{center}
\includegraphics[width=1\linewidth]{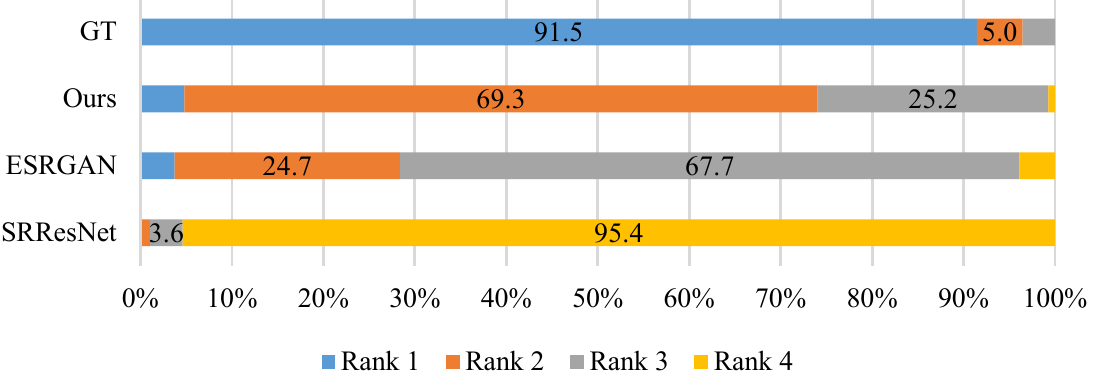} 
\end{center}
   \caption{The ranking results of user studies: SRResNet \cite{ledig2017photo}, ESRGAN \cite{Wang_2018_ECCV_Workshops}, RankSRGAN (ours), and the original HR image.}
\label{fig:user study2}
\end{figure}

As suggested in Figure \ref{fig:user study1}, RankSRGAN has achieved better visual performance against ESRGAN and SRGAN. Since RankSRGAN consists of a base model SRGAN and the proposed Ranker, it can naturally inherit the characteristics of SRGAN and achieve better performance in perceptual metric. Thus, RankSRGAN performs more similar to SRGAN than ESRGAN. Figure \ref{fig:user study2} shows the ranking results of different SR methods. As RankSRGAN has the best performance in perceptual metric, the ranking results of RankSRGAN are second to GT images, but sometimes it even produces images comparable to GT.

\section{Conclusion}
For perceptual super-resolution, we propose RankSRGAN to optimize SR model in the orientation of perceptual metrics. The key idea is introducing a Ranker to learn the behavior  of the perceptual metrics by learning to rank approach. Moreover, our proposed method can combine the strengths of different SR methods and generate better results. Extensive experiments well demonstrate that our RankSRGAN is a flexible framework, which can achieve superiority over state-of-the-art methods in perceptual metric and have the ability to recover more realistic textures. 

\textbf{Acknowledgements}. This work is partially supported by National Natural Science Foundation of China (61876176,  U1613211), Shenzhen Basic Research Program (JCYJ20170818164704758), the Joint Lab of CAS-HK.

{\small
\bibliographystyle{ieee_fullname}
\bibliography{egpaper_final}
}
\newpage



\section*{Supplementary material}

# RankSRGAN: Generative Adversarial Networks with Ranker for Image Super-Resolution
## Supplementary File

Wenlong Zhang[1]    Yihao Liu[1,2]    Chao Dong[1,†]    Yu Qiao[1]
[1]ShenZhen Key Lab of Computer Vision and Pattern Recognition, SIAT-SenseTime Joint Lab,
Shenzhen Institutes of Advanced Technology, Chinese Academy of Sciences, China
[2]University of Chinese Academy of Sciences
{wl.zhang1, yh.liu4, chao.dong, yu.qiao}@siat.ac.cn

## Abstract

*In this supplementary file, we first present more details and additional experimental results of our proposed Ranker. Then, we provide the curves showing the performance of different RankSRGAN models in the ablation study. Finally, we provide more additional qualitative results to compare our networks with the state-of-the-art methods.*

## 1. Details of Ranker

### 1.1. Dataset

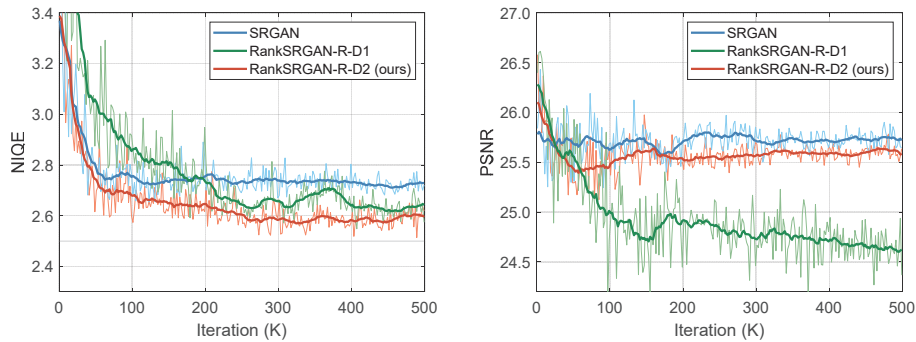

Figure 1. The convergence curves of RankSRGAN with Ranker1 and Ranker2 in NIQE and PSNR.

| Method | Ranker | Dataset | Data Size (k) | SROCC | NIQE | PSNR |
|---|---|---|---|---|---|---|
| RankSRGAN-R-D1 | Ranker1 | DIV2K | 15 | 0.78 | 2.53 | 24.54 |
| RankSRGAN-R-D2 | Ranker2 | DIV2K+Flickr2K | 150 | 0.88 | 2.51 | 25.62 |

Table 1. The performance of RankSRGAN with Ranker1 and Ranker2. R-D1: rank dataset1 (15 K), R-D2: rank dataset2 (150 K)

To analyze the effects of Ranker on RankSRGAN, we use SRResNet [3], SRGAN [3] and ESRGAN [9] to generate two rank datasets with different sizes. We first employ DIV2K [1] to generate rank dataset1 with **15 K** image pairs. Besides, we use DIV2K+Flickr2k [1] to generate rank dataset2 with **150 K** image pairs. Then, we utilize rank dataset1 and rank dataset2 to train Ranker1 and Ranker2, respectively. Finally, the well-trained Ranker1 and Ranker2 are applied on RankSRGAN. Table 1 shows that more data leads to better SROCC, and the Ranker2 with higher SROCC can reach better performance in NIQE and PSNR. The convergence curves are shown in Figure 1.

---

[†]Corresponding author (e-mail: chao.dong@siat.ac.cn)

## 1.2. Network Architecture

The architecture of Ranker is based on the VGG network [8]. We train three VGG networks varying from shallow to deep ones: VGG-8, VGG-12 and VGG-16. Table 2 shows the architecture, the number of parameters, and the performance in different models. Since the VGG-12 can achieve the same accuracy as VGG-16, we apply the VGG-8 and VGG-12 on RankSRGAN. Figure 2 shows the performance of RankSRGAN with different Rankers. The Ranker with higher value of SROCC can achieve better performance when applied on RankSRGAN.

| Model | VGG-8 | VGG-12 (Ours) | VGG-16 |
|---|---|---|---|
| Architecture | Conv3S1-64 | Conv3S1-64 | Conv3S1-64, LReLU |
| | Conv4S2-64, BN, LReLU | Conv4S2-64, BN, LReLU | Conv4S2-64, BN, LReLU |
| | | Conv3S1-128, BN, LReLU | Conv3S1-128, BN, LReLU |
| | | | Conv3S1-128, BN, LReLU |
| | Conv4S2-128, BN, LReLU | Conv4S2-128, BN, LReLU | Conv4S2-128, BN, LReLU |
| | | Conv3S1-256, BN, LReLU | Conv3S1-256, BN, LReLU |
| | | | Conv3S1-256, BN, LReLU |
| | Conv4S2-256, BN, LReLU | Conv4S2-256, BN, LReLU | Conv4S2-256, BN, LReLU |
| | | Conv3S1-512, BN, LReLU | Conv3S1-512, BN, LReLU |
| | | | Conv3S1-512, BN, LReLU |
| | Conv4S2-512, BN, LReLU | Conv4S2-512, BN, LReLU | Conv4S2-512, BN, LReLU |
| | | Conv3S1-512, BN, LReLU | Conv3S1-512, BN, LReLU |
| | | | Conv3S1-512, BN, LReLU |
| | Conv4S2-512, BN, LReLU | Conv4S2-512, BN, LReLU | Conv4S2-512, BN, LReLU |
| | Average pooling | | |
| | FC-100 | | |
| | FC-1 | | |
| Number of params (K) | 7,069 | 13,734 | 19,194 |
| SROCC | 0.83 | 0.88 | 0.88 |

Table 2. The network architecture of Rankers with different depths. The network design draws inspiration from VGG [8] but uses Leaky ReLU activations [5] and strided convolutions instead of pooling layers [7]. Conv3S1-64: Convolutional layer with kernel size 3×3, stride 1 and channel 64. BN: Batch Normalization. LReLU: Leaky ReLU.

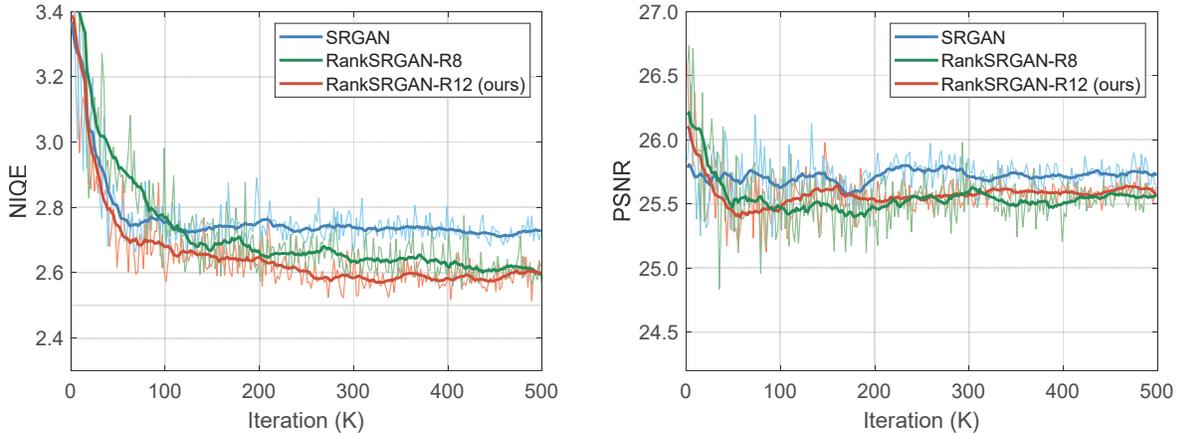

Figure 2. The convergence curves of RankSRGAN with Ranker-VGG-8 and Ranker-VGG-12 in NIQE and PSNR.

## 1.3. Output Distribution

In Table 5 and 6 of the main paper, we quantitatively evaluate the effects of Ranker and "regression" Ranker on RankSR-GAN. To better understand the effects, we provide the histograms of NIQE [6]/Ma [4] label value in the validation dataset of the rank dataset. Furthermore, we plot the histograms of the output scores of different Rankers ("regression" Ranker and our Ranker) in Figure 3. Comparing Figure 3 (b) and (c), Ranker successfully enlarges the distance between SRGAN and ESRGAN. The "regression" Ranker tends to learn the distribution of the NIQE label, while the NIQE values of SRGAN are close ESRGAN. The same observation is also found in Ma metric as shown in Figure 4.

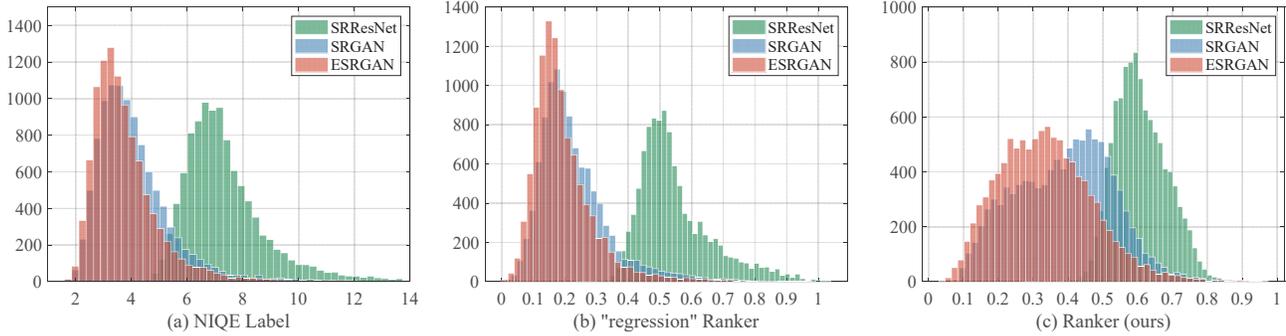

Figure 3. The histograms of (a) NIQE label value, (b) the regression score of "regression" Ranker and (c) the ranking score of Ranker (ours). These graphs illustrate that Ranker (ours) successfully managed to separate the different perceptual levels. (Better view in color version)

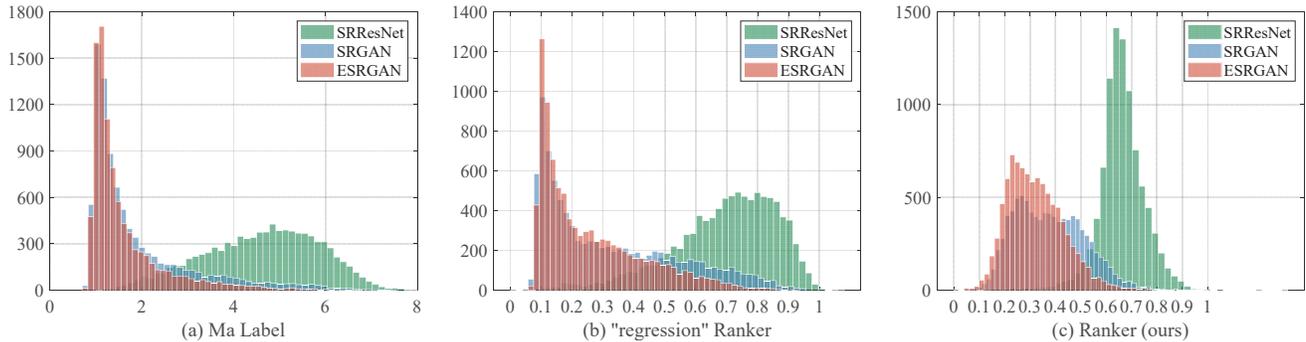

Figure 4. The histograms of (a) Ma label value, (b) the regression score of "regression" Ranker and (c) the ranking score of Ranker (ours). These graphs illustrate that Ranker (ours) successfully managed to separate the different perceptual levels.

## 2. Details of RankSRGAN

### 2.1. Convergence curves for RankSRGAN-(NIQE, Ma, and PI)

As mentioned in the main paper, Ranker can guide the SR model to be optimized in the direction of perceptual metrics. We further present the curves showing that our RankSRGAN can achieve a constant improvement compared with the baseline SRGAN. We provide the curves of RankSRGAN-N, RankSRGAN-M, and RankSRGAN-PI (N: Ranker with NIQE [6], M: Ranker with Ma [4], and PI: Ranker with PI [2]) in Figure 5. Besides, we add the curves of RankSRGAN-N-re and RankSRGAN-M-re (re: "regression" Ranker). We observe that Ranker could help RankSRGAN achieve state-of-the-art performance in the chosen metric. This shows that our method can generalize well on different perceptual metrics. Compared with "regression" Ranker, Ranker can accelerate the convergence of RankSRGAN-N. For RankSRGAN-M, Ranker can still reach state-of-the-art performance (less than 1.40 in ESRGAN), while the "regression" Ranker cannot outperform ESRGAN (1.40).

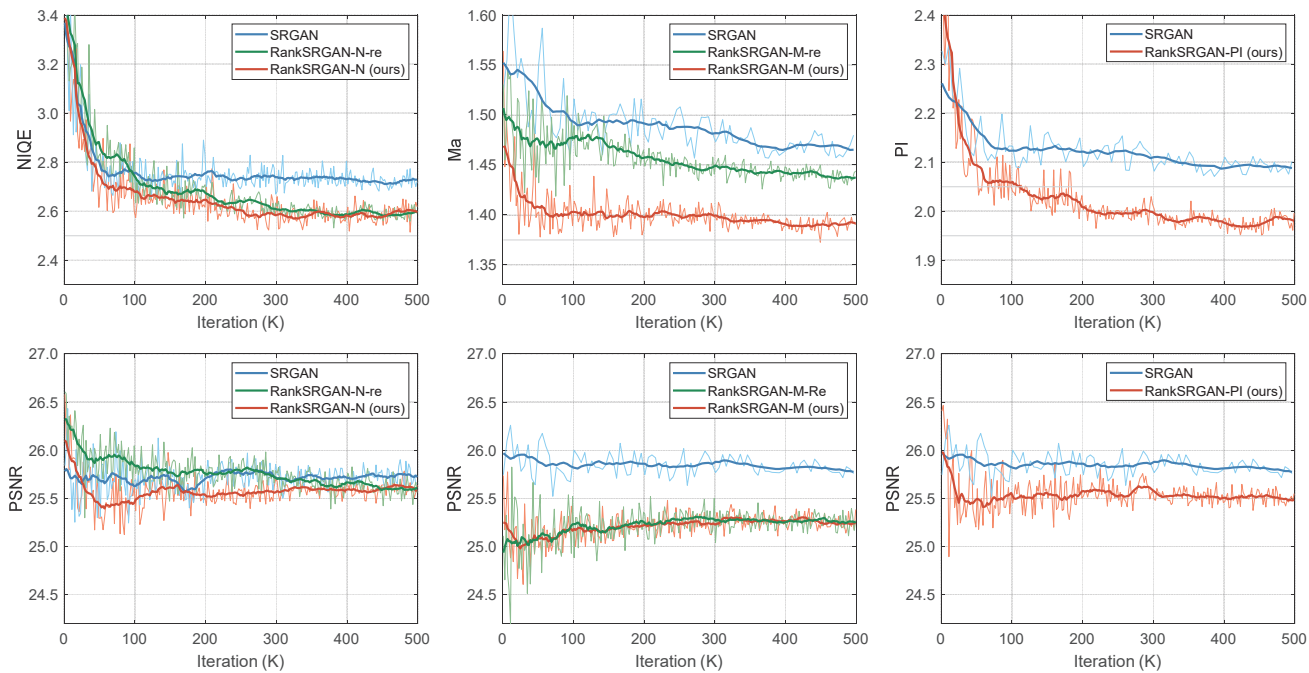

Figure 5. The convergence curves of RankSRGAN$_N$, RankSRGAN$_M$ and RankSRGAN$_{PI}$.

## 2.2. Convergence curves for RankSRGAN-HR

As shown in Figure 6, we present the curves of RankSRGAN and RankSRGAN-HR. To improve the performance of NIQE evaluation, we use (SRResNet, SRGAN, ESRGAN) to generate rank dataset to train Ranker in RankSRGAN. Figure 6 shows that RankSRGAN is consistently better than SRGAN by a large margin. Furthermore, we directly use the ground truth HR to replace ESRGAN . We train our Ranker with the rank dataset (SRResNet, SRGAN, HR) and obtain the new model RankSRGAN-HR. In Figure 6, RankSRGAN-HR achieves better NIQE values than SRGAN. But at the same time, RankSRGAN-HR also constantly improves the PSNR. It achieves a good balance between the perceptual metric and PSNR.

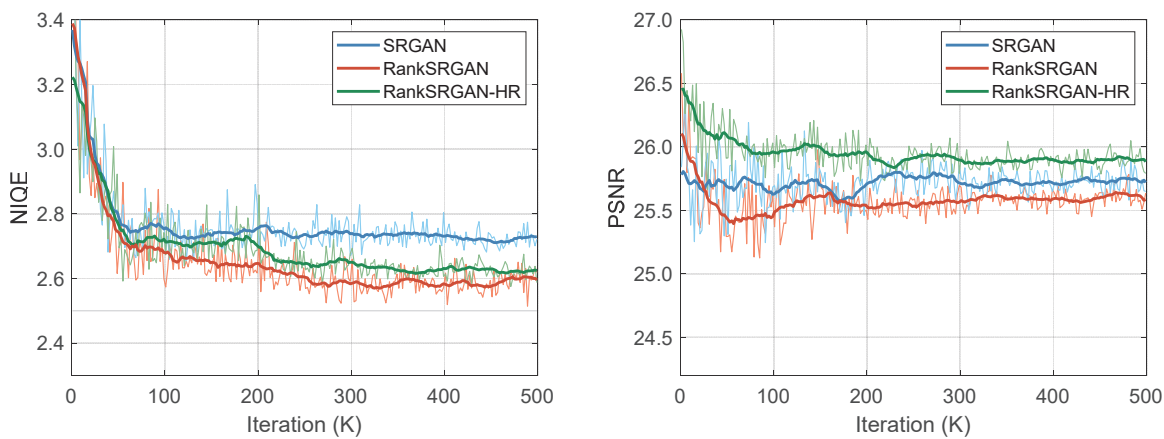

Figure 6. The convergence curves of RankSRGAN-HR in PSNR and NIQE.

## 3. More Qualitative Results

In this section, we provide additional qualitative results (×4 enlargement) to clearly show the effectiveness of our RankSR-GAN. We compare the proposed RankSRGAN with the state-of-the-art perceptual SR methods SRGAN [3] / ESRGAN [9] and PSNR-oriented method SRResNet [3]. We employ NIQE and PSNR to evaluate those SR methods. Lower NIQE value indicates better perceptual quality while higher PSNR indicates that there is less distortion with the Ground-Truth image.

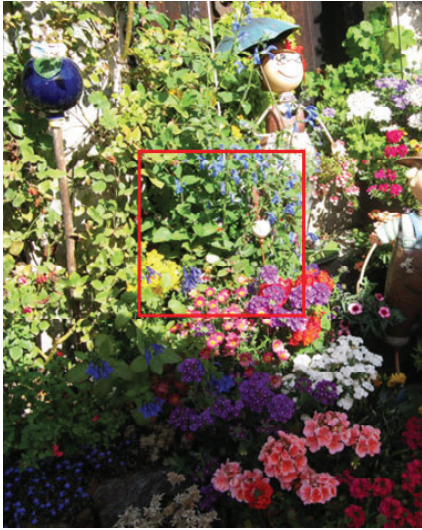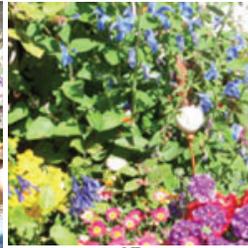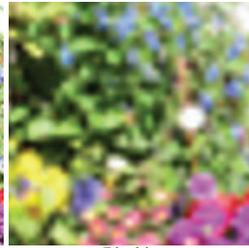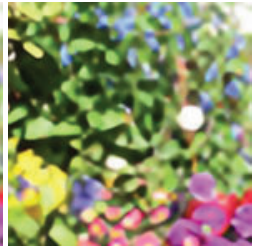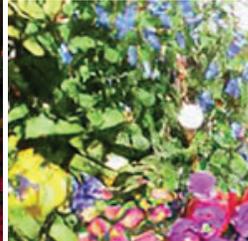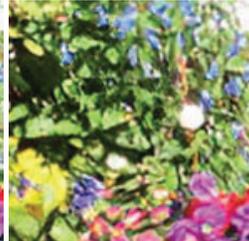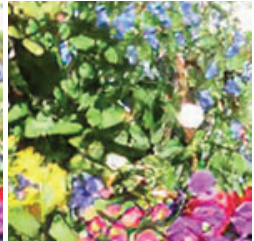

| | GT<br>(NIQE/PSNR) | Bicubic<br>(7.49/20.94) | SRResNet<br>(5.30/21.98) |
| Img 206 | SRGAN<br>(2.33/19.20) | ESRGAN<br>(2.51/19.73) | RankSRGAN (Ours)<br>(**2.17**/19.34) |

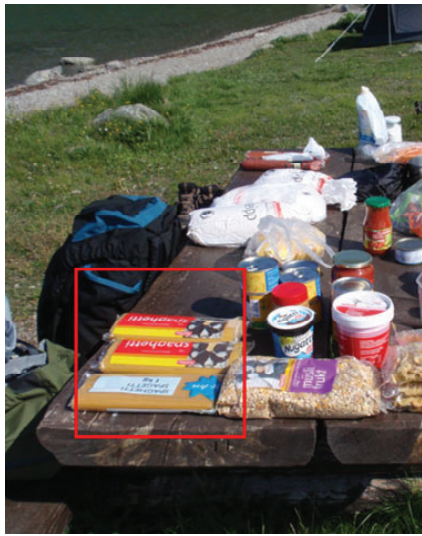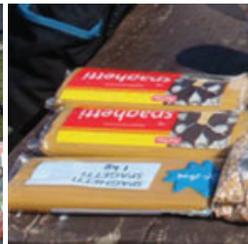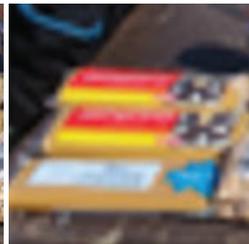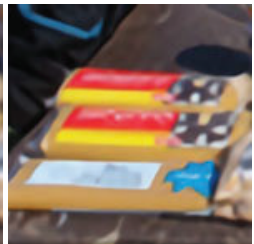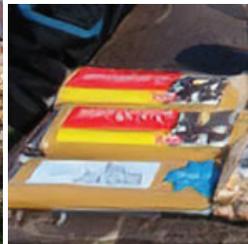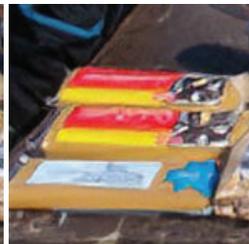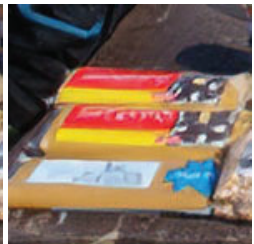

| | GT<br>(NIQE/PSNR) | Bicubic<br>(7.22/26.84) | SRResNet<br>(5.12/28.71) |
| Img 207 | SRGAN<br>(1.99/26.13) | ESRGAN<br>(1.77/25.28) | RankSRGAN (Ours)<br>(**1.77**/26.07) |

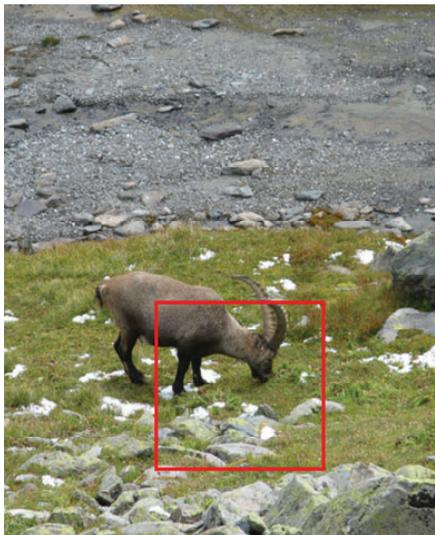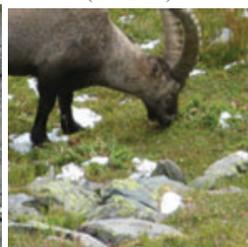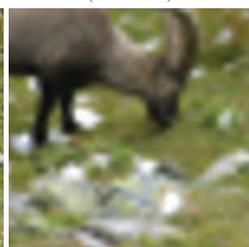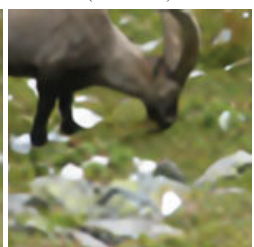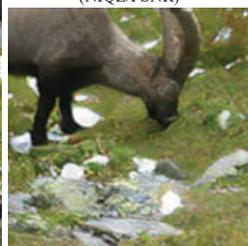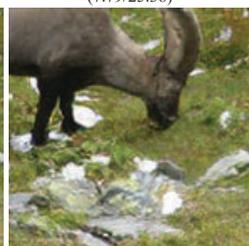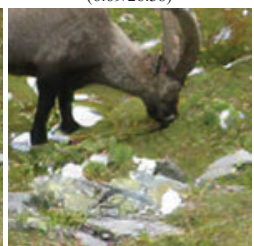

| | GT<br>(NIQE/PSNR) | Bicubic<br>(7.79/25.38) | SRResNet<br>(6.69/26.38) |
| Img 210 | SRGAN<br>(2.70/23.49) | ESRGAN<br>(2.82/23.12) | RankSRGAN (Ours)<br>(**2.66**/23.42) |

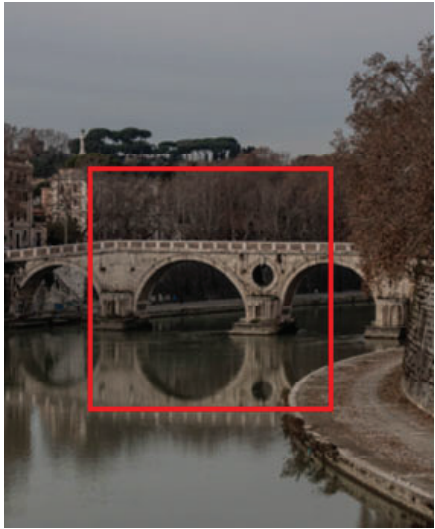 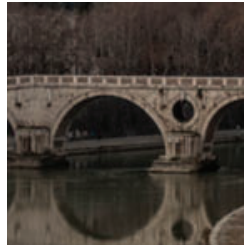 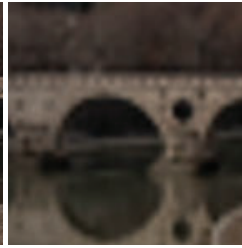 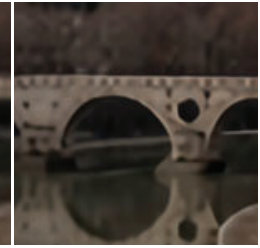
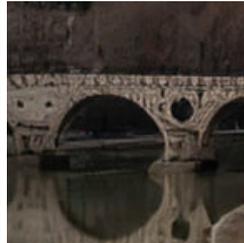 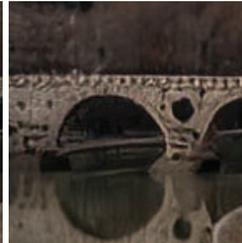 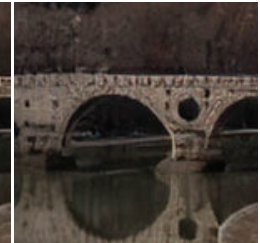

Img 216 | GT (NIQE/PSNR) | Bicubic (7.33/29.25) | SRResNet (5.77/30.27) | SRGAN (2.31/27.52) | ESRGAN (2.53/27.98) | RankSRGAN (Ours) (**2.22**/27.43)

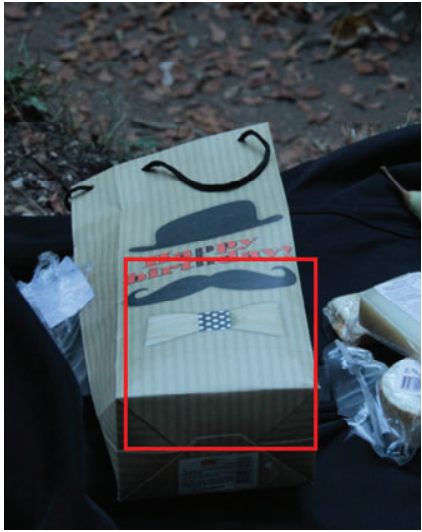 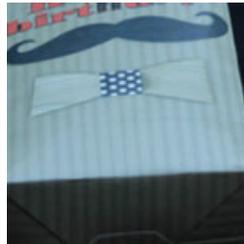 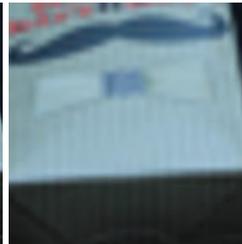 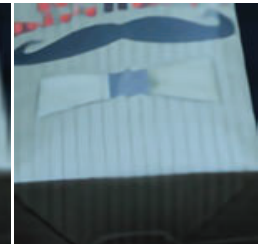
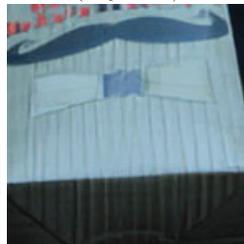 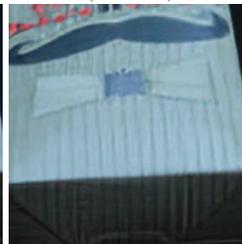 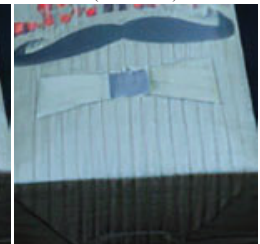

Img 225 | GT (NIQE/PSNR) | Bicubic (7.14/30.27) | SRResNet (5.36/34.42) | SRGAN (2.75/31.88) | ESRGAN (2.47/31.14) | RankSRGAN (Ours) (**2.27**/31.48)

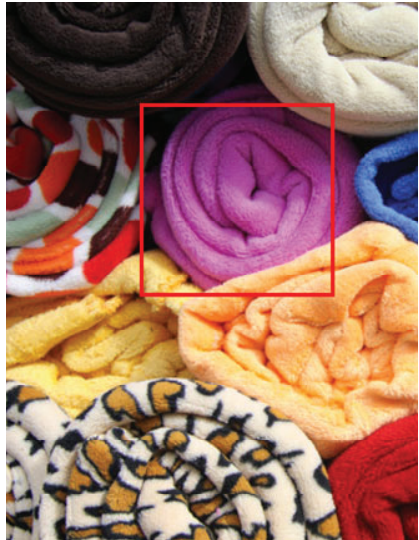 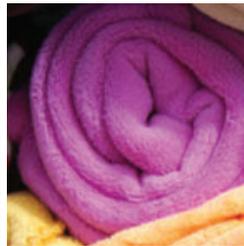 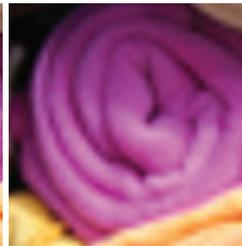 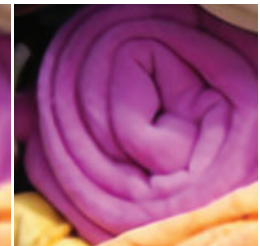
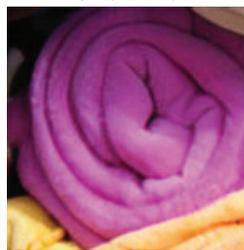 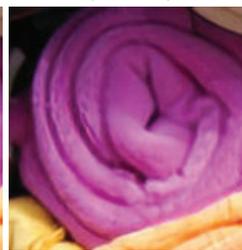 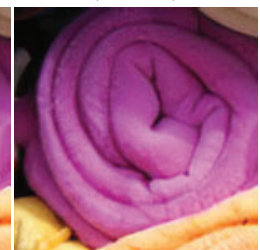

Img 238 | GT (NIQE/PSNR) | Bicubic (7.82/26.51) | SRResNet (6.07/31.00) | SRGAN (3.66/29.40) | ESRGAN (3.49/28.26) | RankSRGAN (Ours) (**3.14**/29.10)

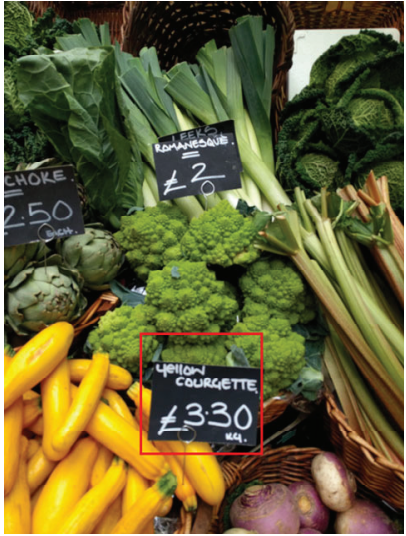 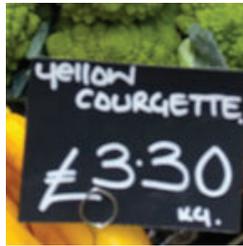 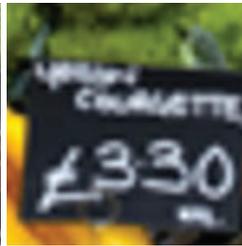 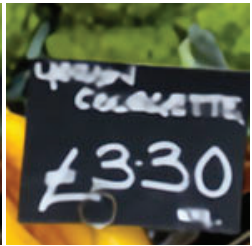
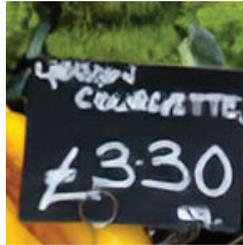 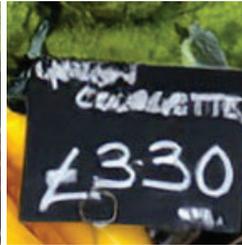 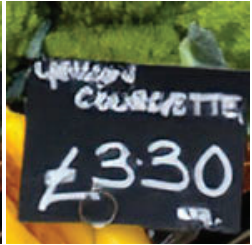

|  | GT<br>(NIQE/PSNR) | Bicubic<br>(7.45/26.37) | SRResNet<br>(5.12/29.83) |
|---|---|---|---|
| Img 243 | SRGAN<br>(2.42/27.38) | ESRGAN<br>(2.32/26.71) | **RankSRGAN (Ours)**<br>(**2.11**/26.96) |

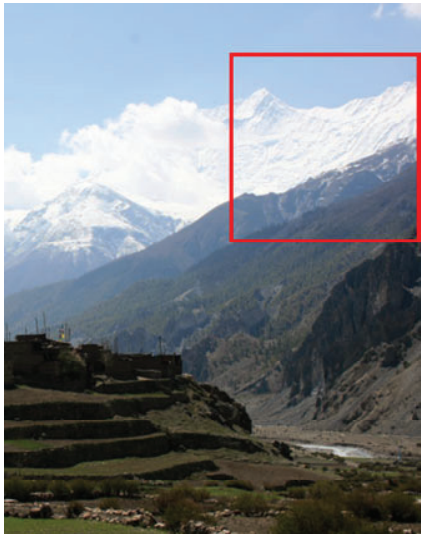 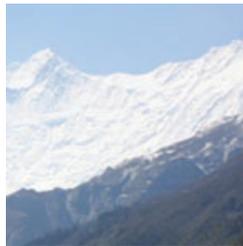 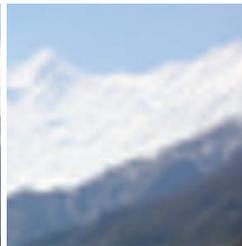 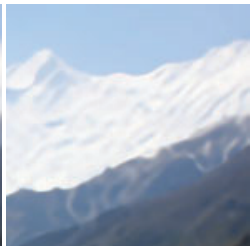
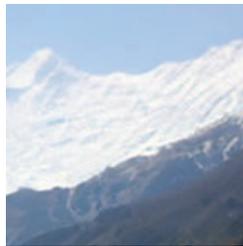 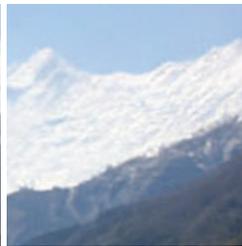 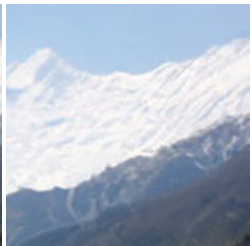

|  | GT<br>(NIQE/PSNR) | Bicubic<br>(7.02/30.60) | SRResNet<br>(6.11/32.64) |
|---|---|---|---|
| Img 246 | SRGAN<br>(2.73/30.32) | ESRGAN<br>(2.91/30.16) | **RankSRGAN (Ours)**<br>(**2.47**/30.29) |

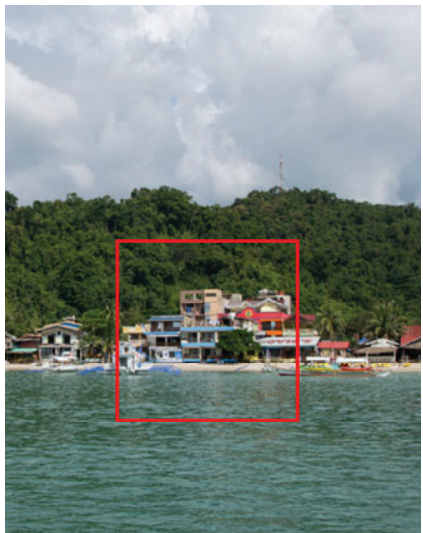 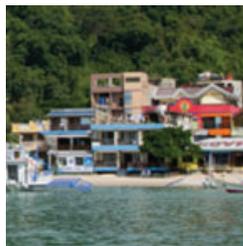 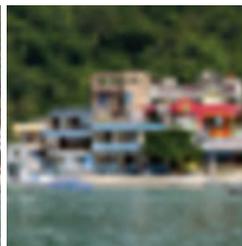 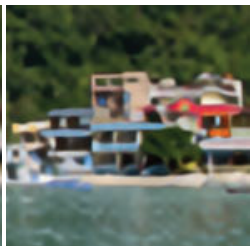
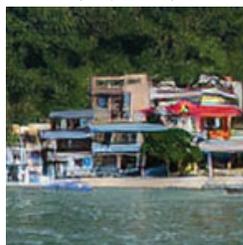 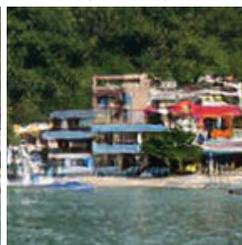 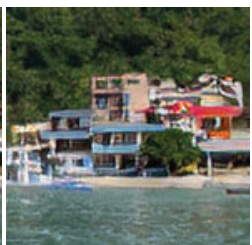

|  | GT<br>(NIQE/PSNR) | Bicubic<br>(7.15/26.83) | SRResNet<br>(6.55/27.69) |
|---|---|---|---|
| Img 256 | SRGAN<br>(2.88/24.18) | ESRGAN<br>(2.68/24.75) | **RankSRGAN (Ours)**<br>(**2.44**/24.85) |



\end{document}